%% file: iclr2020_conference.tex
\documentclass{article} % For LaTeX2e
\usepackage{iclr2020_conference,times}

\input{math_commands.tex}

\usepackage{hyperref}
\usepackage{url}
\usepackage{graphicx}
\usepackage{enumitem}

\title{Teacher-Student chain for efficient semi-supervised histology image classification}

\author{\hspace{2.48cm} Shayne Shaw\textsuperscript{1,2}\thanks{Equal contribution},
\enspace Maciej Pajak\textsuperscript{1}\footnotemark[1],
\enspace Aneta Lisowska\textsuperscript{1}\thanks{Originator of the teacher-student chain idea},\\
\hspace{3.3cm} \textbf{Sotirios A. Tsaftaris\textsuperscript{1,3}},
 \textbf{Alison Q. O’Neil\textsuperscript{1,3}} \\
\hspace{0.55cm} \textsuperscript{1}Canon Medical Research Europe, 
\textsuperscript{2}Heriot-Watt University, 
\textsuperscript{3}University of Edinburgh\\
\hspace{0.3cm}\texttt{\{shayne.shaw, maciej.pajak, aneta.lisowska\}@eu.medical.canon,}\\
\hspace{1.1cm}\texttt{s.tsaftaris@ed.ac.uk, alison.oneil@eu.medical.canon}
}

\iclrfinalcopy % Uncomment for camera-ready version, but NOT for submission.

\begin{document}

\begin{center}
  \maketitle  
\end{center}

% The abstract must be limited to one paragraph.

\begin{abstract}

Deep learning shows great potential for the domain of digital pathology. An automated digital pathology system could serve as a second reader, perform initial triage in large screening studies, or assist in reporting. However, it is expensive to exhaustively annotate large histology image databases, since medical specialists are a scarce resource. In this paper, we apply the semi-supervised teacher-student knowledge distillation technique proposed by \citet{yalniz2019billion} to the task of quantifying prognostic features in colorectal cancer. We obtain accuracy improvements through extending this approach to a chain of students, where each student's predictions are used to train the next student i.e. the student becomes the teacher. Using the chain approach, and only 0.5\% labelled data (the remaining 99.5\% in the unlabelled pool), we match the accuracy of training on 100\% labelled data. At lower percentages of labelled data, similar gains in accuracy are seen, allowing some recovery of accuracy even from a poor initial choice of labelled training set. In conclusion, this approach shows promise for reducing the annotation burden, thus increasing the affordability of automated digital pathology systems.
\end{abstract}

% The abstract must be limited to one paragraph.

%%%%%%%%%%%%%%%%%%%%%%
\section{Introduction}
% motivation

The application of deep learning to computer vision problems shows great potential in the domain of digital pathology \citep{roy2019patch, serag2019translational}. Digital pathology refers to high-resolution scanned images of histology slides containing diseased tissue slices, where the tissue is stained to highlight features of diagnostic value \citep{pantanowitz2018twenty}. Traditionally these slides would be viewed with a light microscope, however, digitisation has created opportunities for automated analysis. A digital pathology machine learning system can serve as a second reader, perform initial triage in large screening studies, or assist in reporting \citep{colling2019artificial}. Since the number of pathologists is low relative to demand \citep{royal2018survey} --- and this gap can be acute in developing countries \citep{wilson2018access} --- such machine learning systems could support continued provision of pathology services now and in the future \citep{salto2019artificial}.

Development of a medical machine learning algorithm can be expensive because it requires specialist annotation, ideally in far greater detail (e.g. pixelwise segmentation) and more comprehensively (e.g. including annotation of specific prognostic features) than would be performed during the usual process of diagnosis. However, medical specialists are a scarce resource, and it may be too difficult or too expensive to exhaustively annotate the large histology image databases at our disposal. Thus, approaches that might enable exploitation of unlabelled data hold great interest.

In this paper we report our extension and application of Yalniz et al.'s semi-supervised teacher-student distillation approach. In this approach, a teacher model is first trained on a small labelled dataset and then applied to a large unlabelled dataset. The student model is then pre-trained on teacher's predictions (pseudo-labels) before fine-tuning on the labelled set. Our contributions are:
\begin{itemize}
    \item Extension of Yalniz et al.'s approach to a chain of teacher-student models, where the student becomes the teacher to the next student, to yield more consistent accuracy improvements, even where the initial labelled training set is suboptimal.
    \item Application of our approach to a digital pathology image classification task, including adaptation of the pseudo-label filtering hyperparameters ($K$, $P$) for this medical imaging setting, to improve the annotation-accuracy trade-off.

\end{itemize}

\subsection{Related work}

% Semi-supervised learning

 Semi-supervised learning is a broad category of machine learning approaches which leverage large quantities of unlabelled data to supplement a small labelled dataset. For example, the additional samples can be used to enforce a model's invariance to augmentations and increase the robustness of the representation \citep{xie2019unsupervised}.
 
 Knowledge distillation was originally introduced to reduce the size of a model with minimal loss of performance  \citep[e.g. see][]{hinton2015distilling}. In this setup the teacher model is used to predict labels for the data and the student model (smaller architecture) is trained using these labels instead of the original labels. However, in the study of \citet{yalniz2019billion} the teacher model predicts labels for another, much larger unlabelled dataset. The student model uses the pseudo-labelled set for pre-training, and the original labelled set for fine-tuning. Unlike in the typical application of knowledge distillation, here the student model's architecture can be the same as the teacher model. The authors were able to boost performance on the ImageNet classification task by leveraging a much larger unlabelled collection of images from social media. In parallel to our work, this technique was extended to a chain of student models by \citep{xie2019self} who additionally found that the introduction of noise to the student training gave performance gains.

%%%%%%%%%%%%%%%%%%%%%%
\section{Methods}

\subsection{Teacher-Student chain approach}

\begin{figure}[!t]
\begin{center}
\includegraphics[width=1\textwidth]{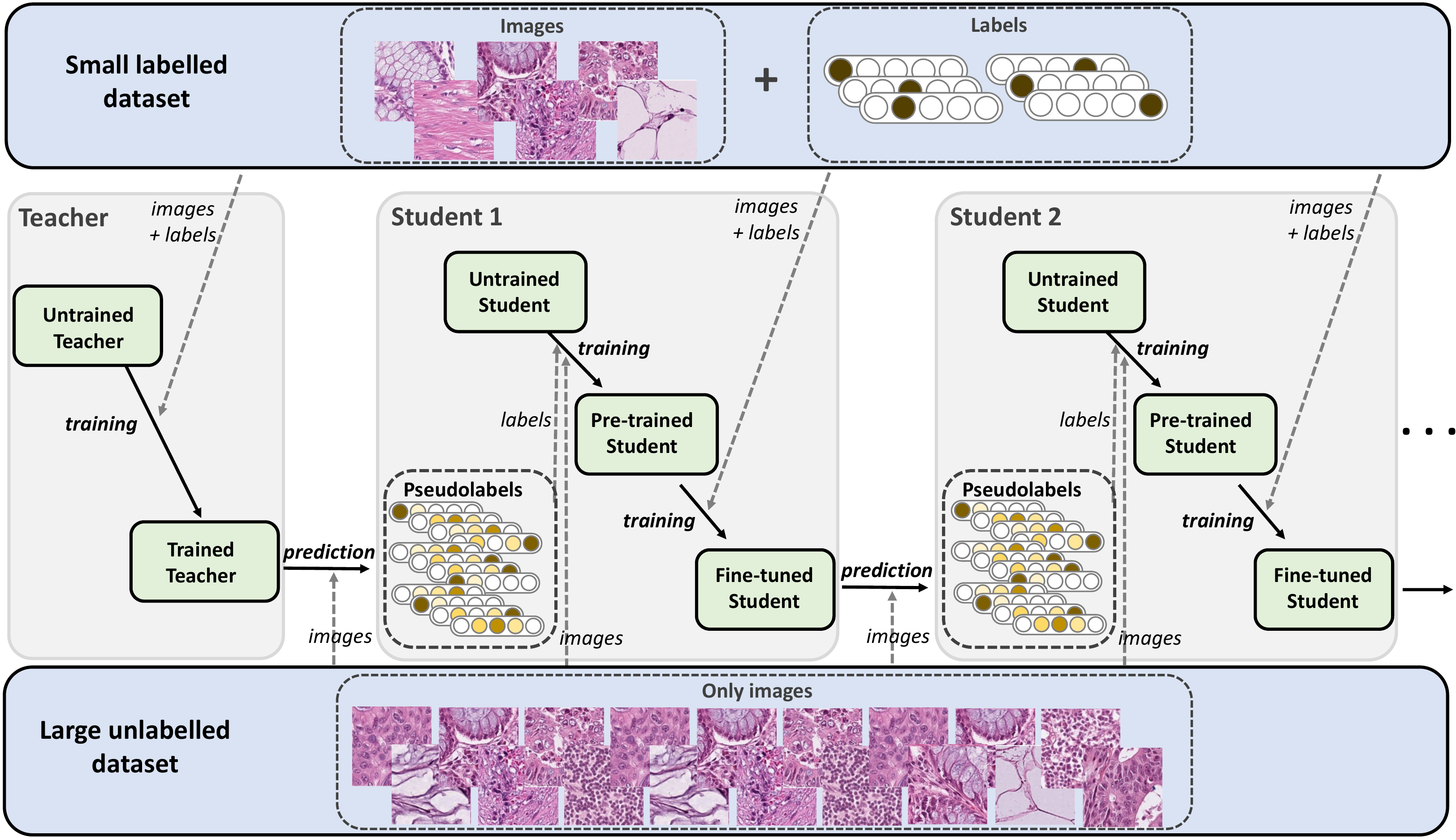}
\end{center}
\caption{Teacher-student chain paradigm for semi-supervised learning. First, the teacher model is trained on the small labelled dataset, and used to generate predictions for the unlabelled dataset. The prediction probabilities are then filtered to retain only the highest probability instances as pseudo-labelled data (soft labels). The student model is trained on the pseudo-labelled data, then fine-tuned on the original labelled dataset. The teacher-student loop is applied iteratively, such that the fine-tuned student generates predictions on the unlabelled dataset and these are the new pseudo-labels used for pre-training of a new student model. This cycle repeats.}
\label{fig:chain}
\end{figure}

Our approach is described in Figure \ref{fig:chain}.

\subsection{Data}
\begin{figure}[htb]
\begin{center}
\includegraphics[width=1.0\textwidth]{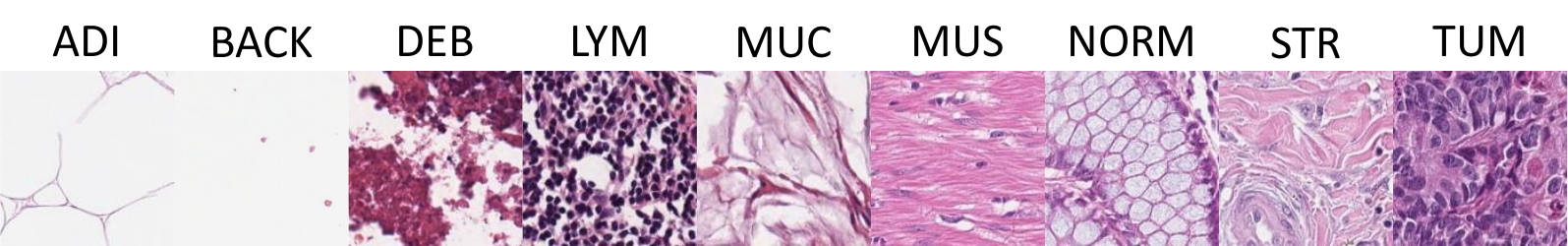}
\end{center}
\caption{Example patches from the dataset used for the classification task \citep{kather2018dataset}.}
\label{fig:data-examples}
\end{figure}

We used a publicly available dataset of 100,000 labelled non-overlapping patches taken from 86 whole-slide images of colorectal tissue samples \citep{kather2018dataset}. The patches are labelled with 9 mutually exclusive categories: adipose (ADI), background (BACK), debris (DEB), lymphocytes (LYM), mucus (MUC), smooth muscle (MUS), normal colon mucosa (NORM), cancer-associated stroma (STR), and colorectal adenocarcinoma epithelium (TUM) --- see Figure~\ref{fig:data-examples} for a sample of patches. \citet{kather2019predicting} showed that the abundance of each of these tissue parts can be combined into a prognostic score. All patches are the same size (224x224 pixels) and are scanned at the same resolution of 0.5$\mu m/px$.

From this dataset, we created a training set of 58,725 randomly selected patches with a balanced distribution of classes, and a validation set of 7,830 patches used for hyperparameter optimisation. We note that since the dataset was provided as patches without a mapping to which of the 86 slides they belong, patches from the same slide may appear in both our training and validation sets. An independent dataset of 7,180 patches from 50 different patients, labelled with the same 9 classes, was made available by Kather et al. as the test set.

\subsection{Single model training procedure}

Following \citet{yalniz2019billion}, we train ResNet-50 \citep{he2016deep} to assign one of the 9 classes to each input patch. The model was trained with the Adam optimiser \citep{kingma2014adam} and categorical cross entropy loss, with a learning rate of $1e-5$ (optimal rate established empirically). The best model weights are selected based on the early stopping set accuracy, the epoch size is kept constant between different experimental conditions. All models are initialised with weights from ImageNet pre-training \citep{deng2009imagenet}, and data are normalised to the same mean ([0.485, 0.456, 0.406]) and standard deviation ([0.229, 0.224, 0.225]) as the ImageNet images used in pre-training to maximise the transfer learning benefit \footnote{https://pytorch.org/docs/stable/torchvision/models.html}.

%%%%%%%%%%%%%%%%%%%%%%
\section{Experiments}
\subsection{Baselines}

\begin{figure}[t!]
\begin{center}
\includegraphics[width=0.48\textwidth]{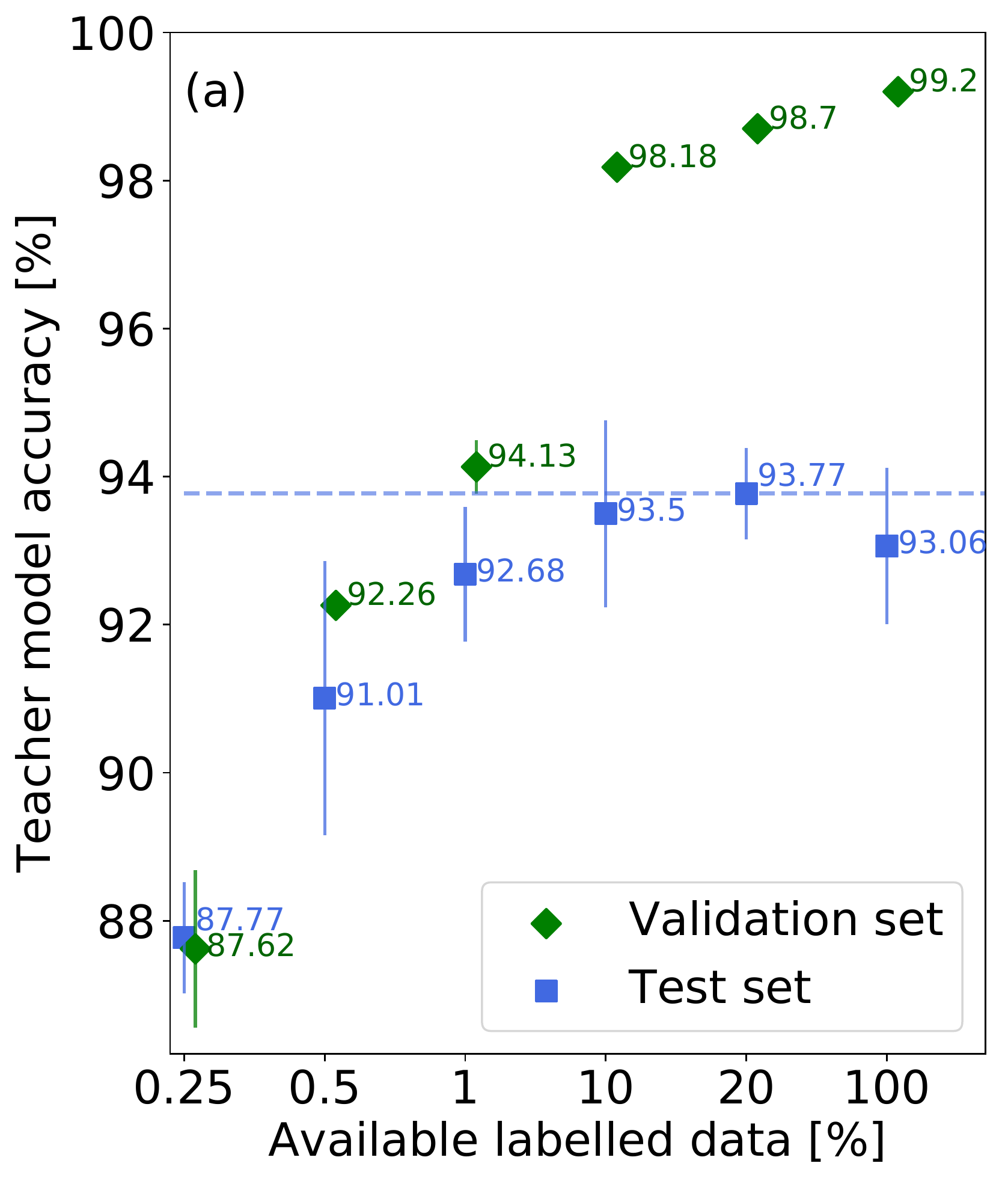}
\includegraphics[width=0.48\textwidth]{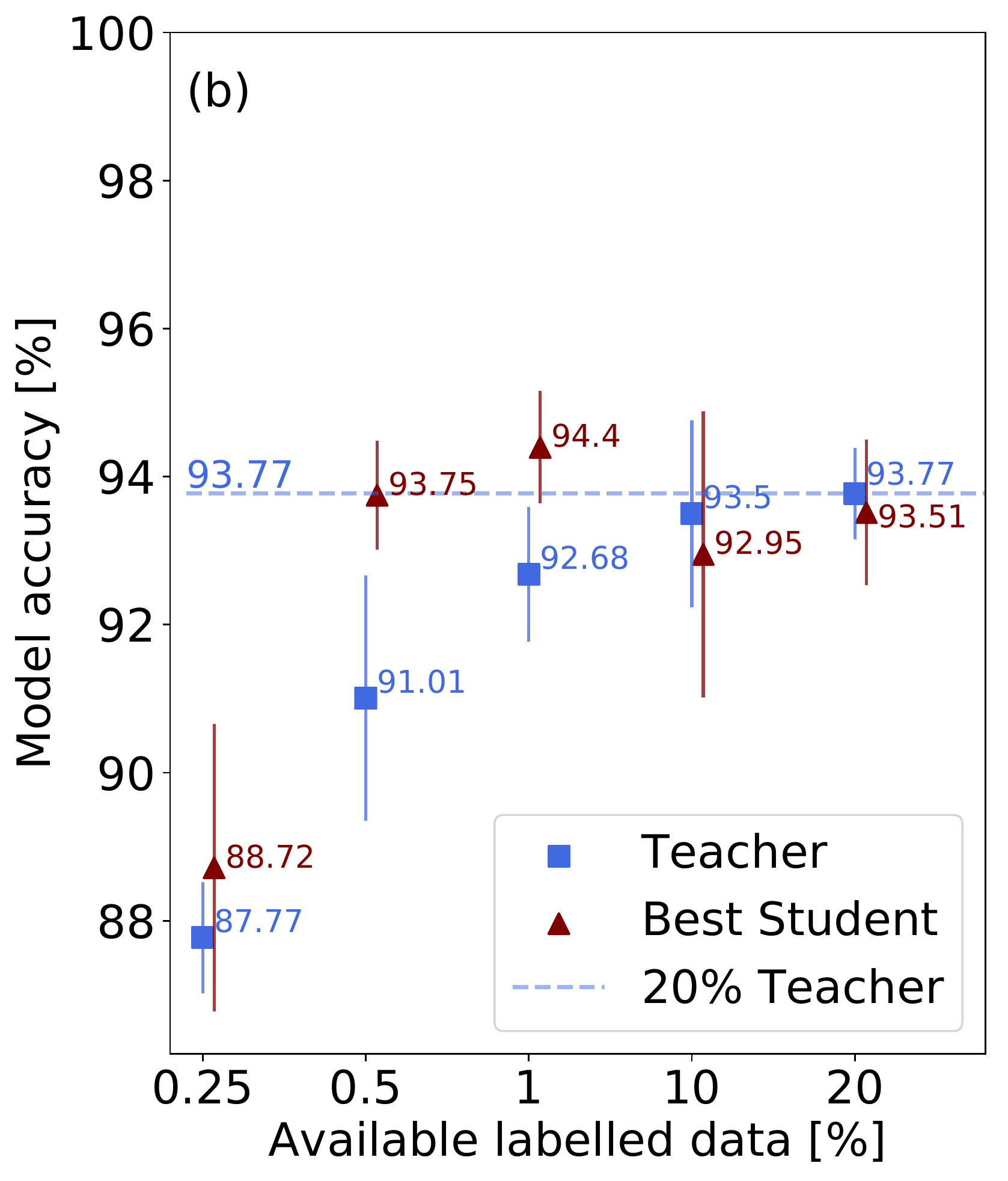}
\end{center}
\caption{Mean and standard deviation of the model performance is shown for: (a) Teacher model trained with different amounts of labelled data (shown on both validation and test sets). (b) Teacher model and best student model for different amounts of labelled data (test set).}
\label{fig:teachers-results}
\end{figure}

We measured the baseline performance of a single teacher model for different sizes of the annotated training set, ranging between 0.25 and 100\%, after first reserving 1\% of the training set as the early stopping set. Patches were selected at random with no enforcement of class balance. We report the mean over 5 experiment runs, where the data splits (early stopping set, labelled training set) are made anew for each run. Results are shown in Figure~\ref{fig:teachers-results}a, where we see that accuracy on the validation set increases from 87.3\% with 0.25\% labelled data up to 99.5\% with 100\% labelled data. However, increasing the amount of labelled data beyond 20\% does not result in further improvement on the test set beyond 93.77\%. We attribute this slightly surprising result to the fact that there is slide resubstitution between our training and validation sets (i.e. patches from the same slides will appear in both sets), and consider this the accuracy ceiling for a teacher model trained on this dataset. It is consistent with the 94.3\% accuracy reported previously by \citet{kather2019predicting} as the best result across different architectures tested in their study.

\subsection{Optimisation of the Teacher-Student parameters}

After the labelled training set is selected, the remaining patches are used as the unlabelled dataset for semi-supervised learning. According to \citet{yalniz2019billion}, the pseudo-labelled dataset can profitably be filtered in two ways, using the $P$ and $K$ parameters:

\hspace{10pt}(1) For each training instance only the top $P$ class probabilities are kept as non-zero.

\hspace{10pt}(2) Only the top $K$ training instances are retained for each class, where instances are classified according to the highest probability class and the highest probability instances are retained.

Optimisation of hyperparameters $K$ and $P$ was performed on a single iteration (first student) of the teacher-student semi-supervised loop, using 1\% labelled data, and the validation set from the same dataset. We found that application of the $P$ parameter made little difference and therefore we chose to retain all class labels. However, use of the $K$ parameter (i.e. only retaining $K$ patches with highest confidence in each class) provided a modest improvement in performance with the optimal value being $K=4000$ per class in the pseudo-labelled training set. $K=4000$ corresponds to approximately 80\% of the available unlabelled training set, i.e. a small proportion of the least certain examples for each class is discarded.

\subsection{Application of the Teacher-Student chain}

\begin{figure}[t!]
\begin{center}
\includegraphics[width=0.32\textwidth]{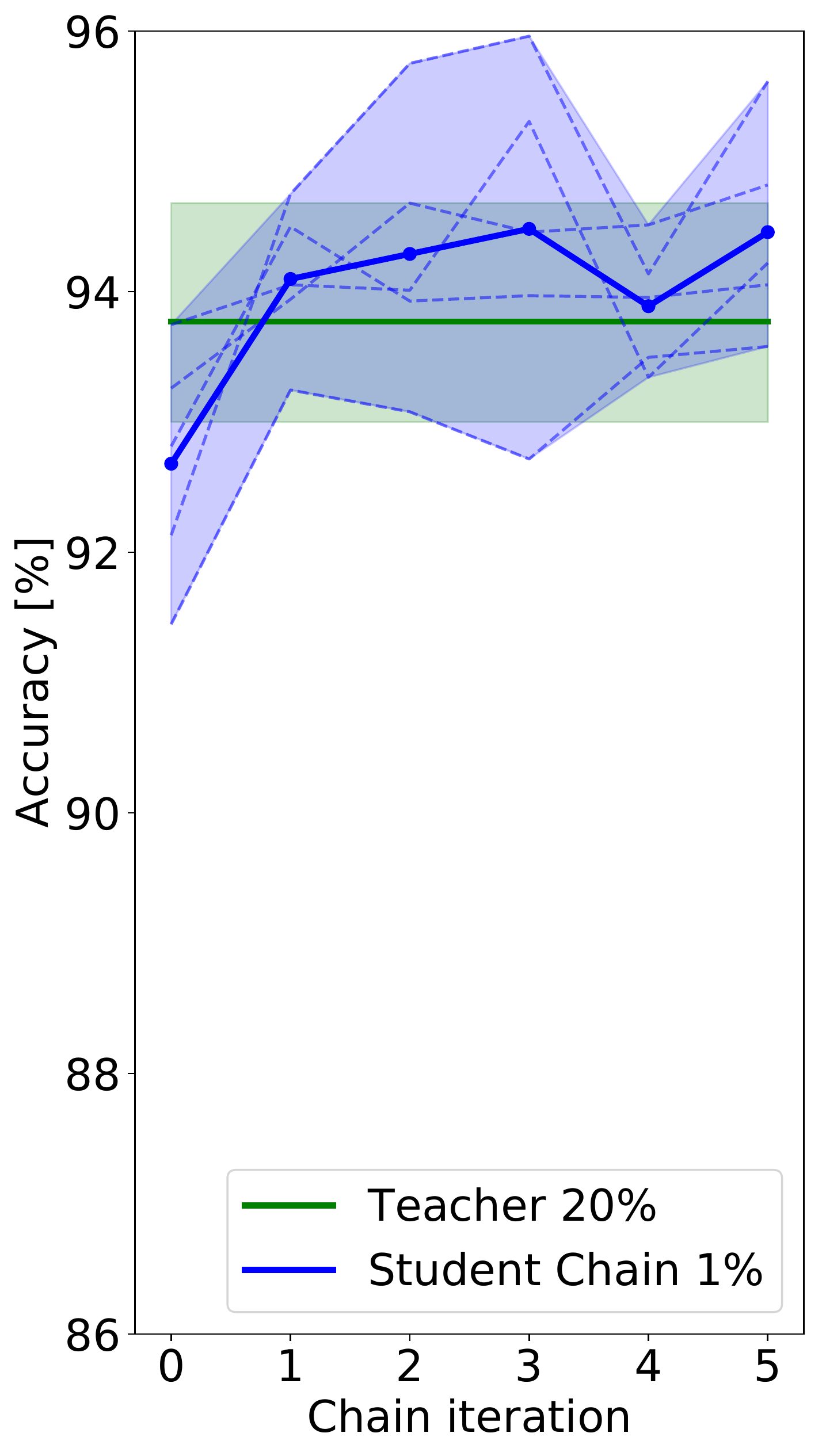}
\includegraphics[width=0.32\textwidth]{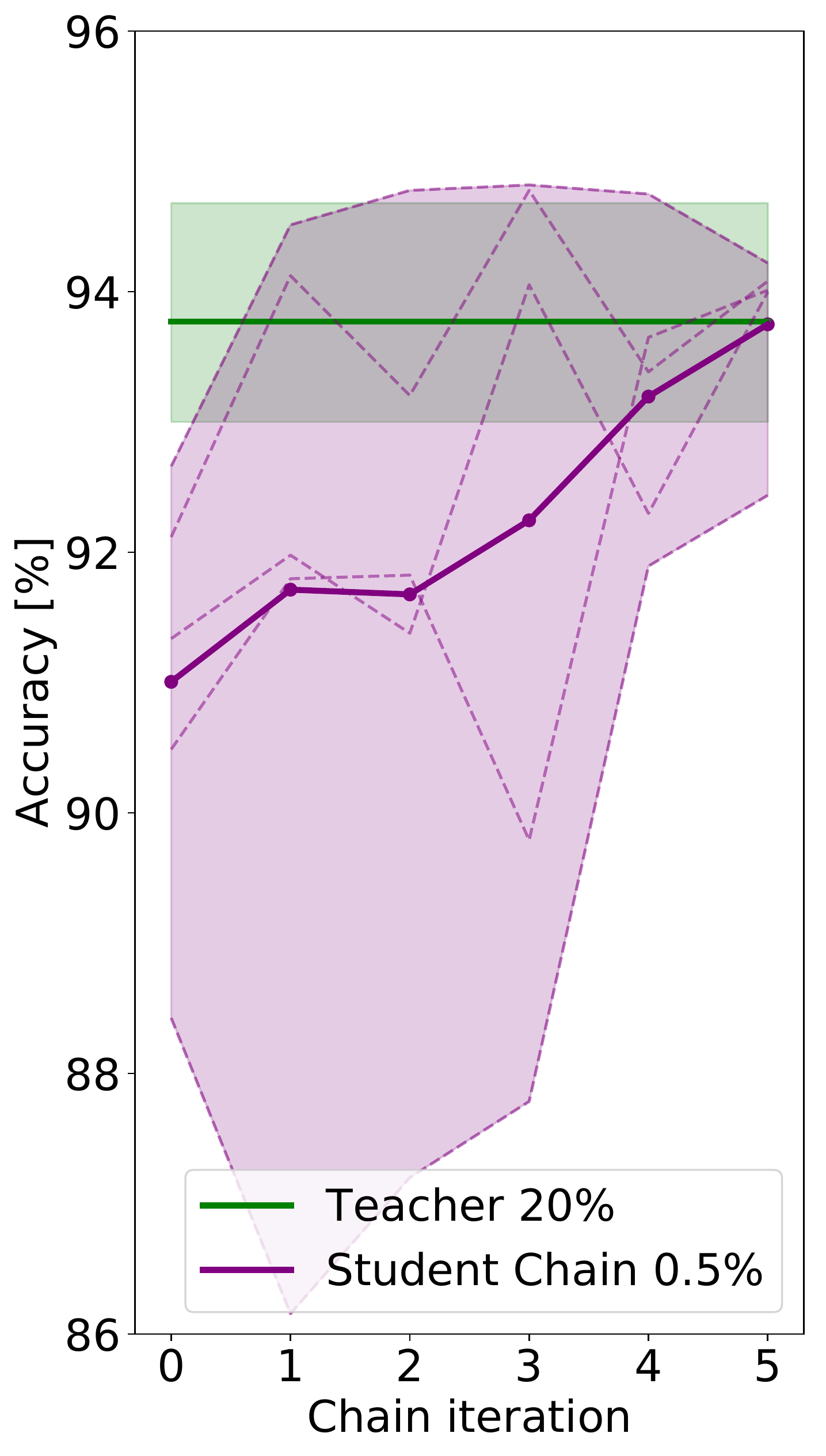}
\includegraphics[width=0.32\textwidth]{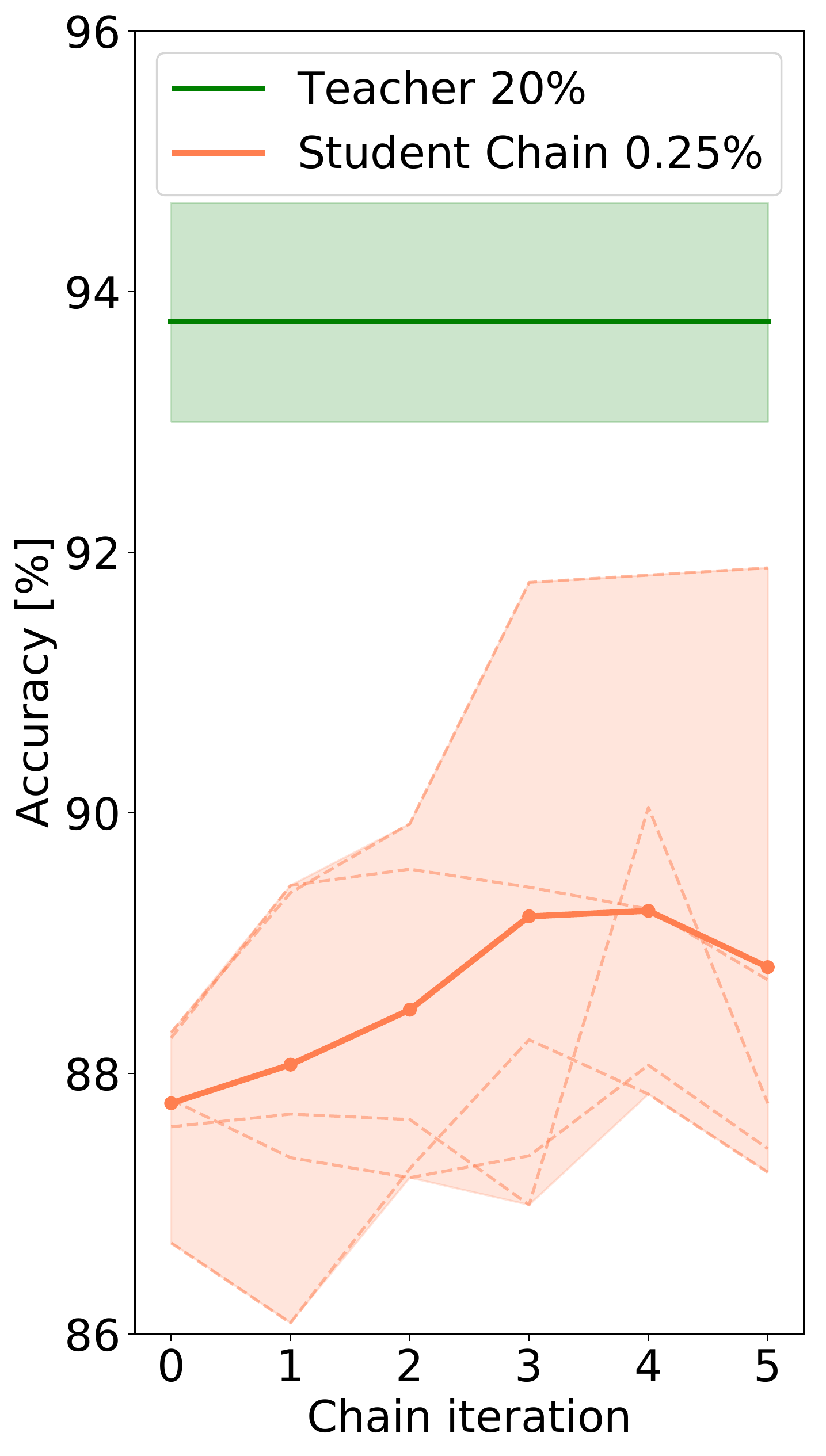}
\end{center}
\caption{Test set results for the semi-supervised teacher-student chain which uses 1\%, 0.5\%, and 0.25\% labelled data. Iteration 0 is the teacher model. The green horizontal line shows the performance of a single model trained with 20\% labelled data (using more data does not further increase the performance). The shaded area indicates the span between the minimum and maximum result across 5 different runs. For the chains, the individual runs are also shown with dashed lines.}
\label{fig:results-chain}
\end{figure}

As previously, a chain of students was trained using a varying amount of labelled data. We observe that the mean accuracy of consecutive student models in the chain improves over at least 3 iterations (see Figure~\ref{fig:results-chain}). The best student trained with 1\% labelled data exceeds the performance of a teacher model (i.e. supervised approach) with 20\% labelled data, and we are able to match the performance of the 20\% teacher model even with only 0.5\% labels (see Figure~\ref{fig:teachers-results}b).

However, we observe large variance and although on average, the performance for subsequent students increases, there is no consistency in the iteration where the best result was achieved across multiple runs. This variance increases further as we decrease from 1\% of data to smaller amounts of labelled data (see results for 1\%, 0.5\% and 0.25\% in Figure~\ref{fig:results-chain}). It may be that better models might emerge with longer chains than 5, however since we rely on scientific rigour to avoid overfitting (e.g. choosing the best model based on the validation set rather than the test set), longer chains are probably best avoided. This interestingly parallels the common experience of the machine learning scientist when training a model using backpropagation, that careful judgement of the stopping point must be applied to find the best parameters that neither overfit not underfit the true distribution.

Debris, Muscle, Normal, and Stroma categories were consistently the most difficult to distinguish for the teacher model. Stroma remained a difficult category for the best student in the chain but differentiation between the other three improved significantly.

%%%%%%%%%%%%%%%%%%%%%%
\section{Discussion and Conclusions}

We have shown that iterative application of the teacher-student semi-supervised approach with a chain of student models, using a small labelled set supplemented by a large unlabelled set of data, results in improvements to the accuracy of the student models compared to supervised training only on the small labelled dataset. Our approach therefore offers a route to reducing the annotation burden, increasing the affordability of developing machine learning solutions for tasks in the domain of digital pathology. We caution that our numbers (e.g. 0.5\% provides the performance of 100\%) are favourable since we took a limited training sample across many patients --- which in practice requires significant annotation of many slides --- in contrast to the real-world scenario where we would be more likely to label a few slides and have less diversity present in our training set.

Finally, we note that it is satisfying that the performance improvements from our semi-supervised teacher-student chain approach and from the choice of labelled training set are cumulative. Further work on the strategy for choosing the best set of examples to label from a large unlabelled dataset could complement the findings of this paper, to give consistently high performance.

% optional

% optional

\clearpage
\bibliography{iclr2020_conference}
\bibliographystyle{iclr2020_conference}

% \appendix

\end{document}

%% file: math_commands.tex
\usepackage{amsmath,amsfonts,bm}

\def\eqref#1{equation~\ref{#1}}
\def\1{\bm{1}}

\DeclareMathAlphabet{\mathsfit}{\encodingdefault}{\sfdefault}{m}{sl}
\SetMathAlphabet{\mathsfit}{bold}{\encodingdefault}{\sfdefault}{bx}{n}